\documentclass[amsmath,amssymb,superscriptaddress,nobalancelastpage,prb,twocolumn]{revtex4-2} 

\AtBeginDocument{%
\newwrite\bibnotes
\def\bibnotesext{Notes.bib}
\immediate\openout\bibnotes=\jobname\bibnotesext
\immediate\write\bibnotes{@CONTROL{REVTEX41Control}}
\immediate\write\bibnotes{@CONTROL{%
apsrev41Control,author="08",editor="1",pages="1",title="0",year="1"}}
\if@filesw
\immediate\write\@auxout{\string\citation{apsrev41Control}}%
\fi
}%

\usepackage{graphicx}
\usepackage{varioref}
\usepackage{xr-hyper}
\usepackage{xcolor}
\usepackage{hyperref}
\hypersetup{colorlinks,linkcolor=blue,urlcolor=blue,citecolor=blue}

\usepackage{ulem}
\usepackage{braket}
\usepackage{array}
\usepackage{float}
\usepackage{orcidlink}
\usepackage{multirow}
\usepackage{tabularx}
\newcommand{\STO}{SrTiO$_3$}
\newcommand{\LSNO}{La$_{1.5}$Sr$_{0.5}$NiO$_4$}
\newcommand{\CVS}{CsV$_3$Sb$_5$}

\newcolumntype{M}[1]{>{\centering\arraybackslash}m{#1\textwidth}}

\begin{document}

\title{Autonomous Diffractometry Enabled by Visual Reinforcement Learning}

\author{J.~Oppliger\textbf{~\orcidlink{0000-0002-0712-4343}}}
\email{jens.oppliger@physik.uzh.ch}
\affiliation{Physik-Institut, Universit\"{a}t Z\"{u}rich, Winterthurerstrasse 190, CH-8057 Z\"{u}rich, Switzerland}

\author{M.~Stifter}
\affiliation{Physik-Institut, Universit\"{a}t Z\"{u}rich, Winterthurerstrasse 190, CH-8057 Z\"{u}rich, Switzerland}

\author{A.~R\"{u}egg}
\affiliation{Physik-Institut, Universit\"{a}t Z\"{u}rich, Winterthurerstrasse 190, CH-8057 Z\"{u}rich, Switzerland}

\author{I.~Biało~\orcidlink{0000-0003-3431-6102}}
\affiliation{Physik-Institut, Universit\"{a}t Z\"{u}rich, Winterthurerstrasse 190, CH-8057 Z\"{u}rich, Switzerland}

\author{L.~Martinelli~\orcidlink{0000-0003-4978-8006}}
\affiliation{Physik-Institut, Universit\"{a}t Z\"{u}rich, Winterthurerstrasse 190, CH-8057 Z\"{u}rich, Switzerland}

\author{P.~G.~Freeman}
\affiliation{Jeremiah Horrocks Institute for Mathematics, Physics and Astronomy, University of Central Lancashire, Preston PR1 2HE, United Kingdom}

\author{D.~Prabhakaran}
\affiliation{Department of Physics, Clarendon Laboratory, University of Oxford, Oxford OX1 RPU, United Kingdom}

\author{J.~Zhao}
\affiliation{State Key Laboratory of Surface Physics and Department of Physics, Fudan University, Shanghai 200433, China}

\author{Q.~Wang~\orcidlink{0000-0002-8741-7559}}
\affiliation{Department of Physics, The Chinese University of Hong Kong, Shatin, Hong Kong, China}
\affiliation{State Key Laboratory of Quantum Information Technologies and Materials, The Chinese University of Hong Kong, Shatin, Hong Kong, China}

\author{J.~Chang~\orcidlink{0000-0002-4655-1516}}
\email{johan.chang@physik.uzh.ch}
\affiliation{Physik-Institut, Universit\"{a}t Z\"{u}rich, Winterthurerstrasse 190, CH-8057 Z\"{u}rich, Switzerland}

\begin{abstract}
\textbf{Automation underpins progress across scientific and industrial disciplines. Yet, automating tasks requiring interpretation of abstract visual information remain challenging. For example, crystal alignment strongly relies on humans with the ability to comprehend diffraction patterns. Here we introduce an autonomous system that aligns single crystals without access to crystallography and diffraction theory. Using a model-free reinforcement learning framework, an agent learns to identify and navigate towards high-symmetry orientations directly from Laue diffraction patterns. Despite the absence of human supervision, the agent develops human-like strategies to achieve time-efficient alignment across different crystal symmetry classes. With this, we provide a computational framework for intelligent diffractometers. As such, our approach advances the development of automated experimental workflows in materials science.}
\end{abstract}

\maketitle

\begin{figure*}
\centering
\includegraphics[width=0.95\textwidth]{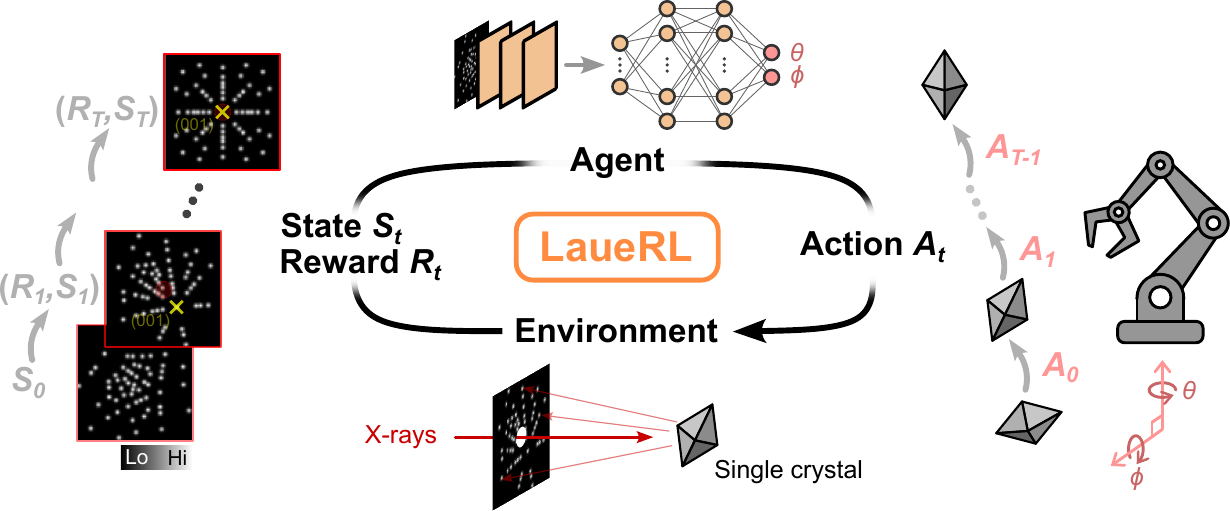}
\caption{\textbf{Schematic of agent-environment interaction for Laue single crystal alignment.} The environment consists of an x-ray Laue back-reflection setup yielding a two-dimensional diffraction pattern (state $S_t$, linear false color scale), here shown for a cubic crystal structure. Given $S_t$, an agent --- described by a deep neural network --- is trained to predict an action $A_t$ consisting of two rotation angles $(\theta,\phi)$ around two perpendicular axes, executed by a robotic arm. This results in a reward signal $R_{t+1}$ and a new state $S_{t+1}$. The state-action-reward loop ends when a predefined high-symmetry orientation --- here chosen as (001) (yellow crosses) --- is achieved within an angular tolerance (red circles) after $T$ steps.}
\label{fig:figure_1_schematic}
\end{figure*}

Learning guides evolution by shaping which behaviors get selected, without 
those behaviors 
being genetically hardwired~\cite{schultz_neural_1997,hinton_how_1996}. 
This suggests that a system can learn adaptive behaviors from experience alone, without  programmed rules or a  physical model of its environment. 
Additionally, it has been proposed that learning of reward-maximizing behaviors alone may suffice for general intelligence~\cite{silver_reward_2021}.
As such, 
reinforcement learning~(RL)~\cite{sutton_reinforcement_2018} and achieving optimal control from raw sensory inputs, combined with evolutionary strategies~\cite{stafylopatis_autonomous_1998,salimans_evolution_2017,majid_deep_2024}, is 
regarded as central to the pursuit of  
artificial general intelligence~\cite{mnih_humanlevel_2015,silver_mastering_2016,silver_mastering_2017,jaderberg_humanlevel_2019,schrittwieser_mastering_2020}. At its core, RL can be 
modeled as 
a Markov decision process~(MDP)~\cite{bellman_markovian_1957}, which captures sequential decision-making through the interaction between a learning agent and its environment.
In the past decades, reinforcement learning has witnessed substantial progress, delivering 
increasingly efficient algorithms capable of solving problems of growing complexity~\cite{watkins_qlearning_1992,silver_deterministic_2014,schulman_proximal_2017,lillicrap_continuous_2015,haarnoja_soft_2019}.\\

\noindent
Traditionally, RL agents are trained using state-based observations, which satisfy the Markov property. That is, they fully describe the past agent-environment interaction to make an informed decision about the future.  
When such compact state representations are unavailable, however, agents must instead learn directly from high-dimensional raw sensory inputs such as pixels. In this setting, model-free methods, which learn a control policy purely through trial-and-error without an explicit model of the environment, are known to suffer from severe sample inefficiency~\cite{mnih_humanlevel_2015,tassa_deepmind_2018}. This is because the agent must simultaneously learn a useful representation of its observations and an effective policy. 


\noindent
Remarkably, humans and animals in general excel at the task of processing and interpreting visual inputs, raising the question as to whether artificial intelligence can 
achieve similar performance~\cite{lake_building_2017}.
As such, visual RL has recently seen significant developments~\cite{hafner_learning_2019,laskin_curl_2020,yarats_image_2020,yarats_mastering_2021,cetin_stabilizing_2022,yuan_pretrained_2022,xu_drm_2023,hafner_mastering_2025},
substantially increasing its sample-efficiency and making it possible to achieve state-of-the-art results compared to conventional state-based RL. Extending these methods from simulation to physical systems naturally motivates their application in robotic control, where real-time feedback from raw sensory inputs and sequential decision-making are central requirements for dynamic object manipulation~\cite{zhang_visionbased_2015,kalashnikov_scalable_2018}. The resulting capabilities are particularly compelling for repetitive tasks that require accuracy and repeatability. 
One of such tasks is a common occurrence in condensed matter research. Studies of structural, electronic, and magnetic properties of materials with advanced scattering techniques often rely on the alignment of single crystals~\cite{kuspert_engineering_2024, simeth_microscopic_2023,ronning_electronic_2017}.
Insights gained at this level contribute to the conceptual foundation of materials research, informing the design principles of functional materials.\\

\noindent
White-beam x-ray diffraction --- so-called Laue diffraction (see \mbox{Supplementary~Note~1}) --- is a 
commonly used technique for crystallographic alignment.
Laue diffraction patterns provide an abstract visual representation of the crystals lattice structure in reciprocal space. Aligning a single crystal along specific crystallographic high-symmetry directions~\cite{bialo_magnetic_2025} usually relies on an experienced human 
to navigate in reciprocal space. Experiments requiring the simultaneous alignment of dozens or hundreds individual single crystals such as inelastic neutron scattering~\cite{shen_evidence_2016,song_robust_2016,simeth_microscopic_2023} --- invaluable to the study of structural and magnetic properties of materials --- are therefore notoriously dependent on 
human labor.\\


\noindent
In this work, we present \textit{LaueRL}, a novel approach to solve the single crystal alignment problem by means of visual model-free reinforcement learning. In particular, we highlight the training of a Laue agent solely based on simulated diffraction patterns for different crystal systems. The  agent autonomously learns to align the crystals along specific high-symmetry directions, leveraging high-symmetry lines as reference features during navigation. 
Simple training randomization techniques enable 
robust transfer of learned behaviors to real-world experimental settings. As such, the Laue agent can independently operate an x-ray Laue instrument and autonomously align single crystals.




\begin{figure*}
\centering
\includegraphics[width=0.95\textwidth]{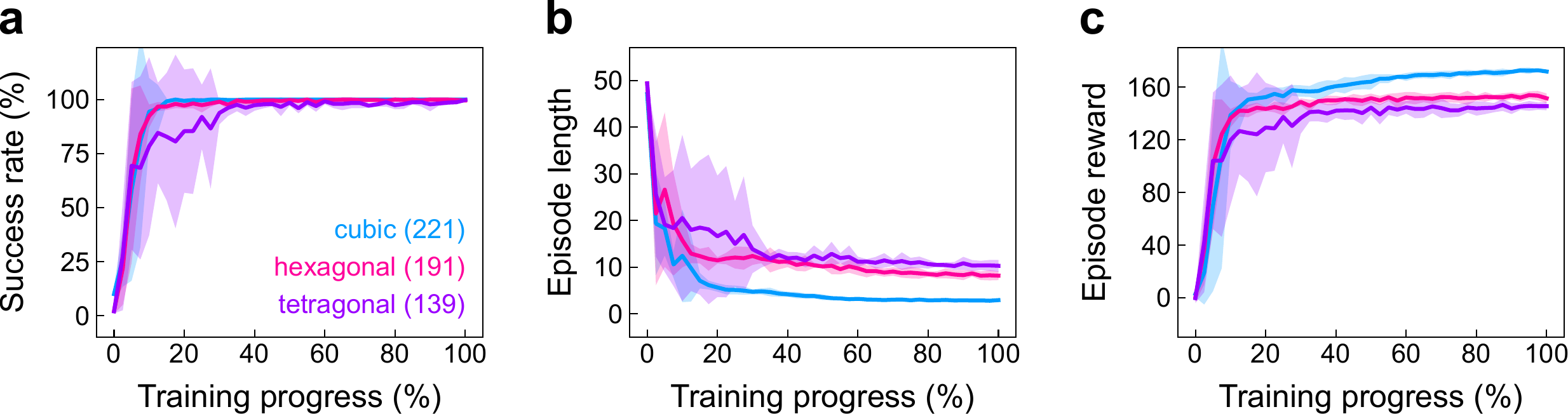}
\caption{\textbf{Agent training curves for different crystal structures.} (a-c) Success rate, episode length, and episode reward as a function of percentage training cycles for selected crystal structures (space groups) as indicated. Solid lines represent the mean, while shaded areas indicate the 95\% confidence interval calculated from five training runs with different random seeds.}
\label{fig:figure_2_training_curves}
\end{figure*}

\subsection*{Environment design and reinforcement learning}
\noindent
A schematic illustration of the agent-environment interaction for Laue single crystal alignment is shown in \mbox{Figure~\ref{fig:figure_1_schematic}}. The environment 
models 
Laue x-ray diffraction 
in the back-reflection geometry 
(see Methods section). Single crystal Laue patterns 
are composed of a set of 
diffraction spots, which depends on the crystallographic space group, cell parameters, atomic basis 
and crystal orientation. 
The  task of the Laue agent is to locate a principal crystallographic axis, within a defined angular tolerance.
For demonstration purposes, we show the identification of the $c$-axis of a cubic 
crystal in \mbox{Figure~\ref{fig:figure_1_schematic}}.
An initial Laue pattern from an  arbitrarily oriented crystal, 
forms the starting state $S_0$ of the alignment process. A non-linear function approximator (agent) 
receives the current state and predicts an action $A_t=(\theta_t,\phi_t)$, defined by a crystal rotation around two perpendicular axes.
The action $A_t$ is then executed by a robotic arm, yielding 
 a new state $S_{t+1}$ and reward $R_{t+1}$.
The reward 
scales with 
inverse angular distance to the high-symmetry point and incentives 
the agent to align the crystal 
with a minimum number steps (see Methods section). 
This state-action-reward sequence 
repeats until a final state $S_T$ is reached. 
Finally, we denote a single alignment episode as the sequence $\left\lbrace (S_0,A_0),(R_1,S_1,A_1),...(R_T,S_T) \right\rbrace$ with $T \in \mathbb{N}$.\\

\noindent
Our implemented solution follows the paradigm of model-free off-policy learning based on actor-critic methods~\cite{lillicrap_continuous_2015,haarnoja_soft_2019,xu_drm_2023}. 
The actor network is comprised of an encoder, which extracts features from the raw Laue pattern using a small convolutional neural network~(CNN), followed by a fully-connected network (multilayer perceptron,~MLP), which uses a latent representation of the encoded features to predict an action (see top part of schematic in Figure~\ref{fig:figure_1_schematic}). The predicted action is 
evaluated by a double-critic network~\cite{hasselt_deep_2016,fujimoto_addressing_2018}, which is constructed identically to the MLP of the actor network. More information about the RL training including hyperparameters can be found in the Methods section.

\begin{figure*}
\centering
\includegraphics[width=0.95\textwidth]{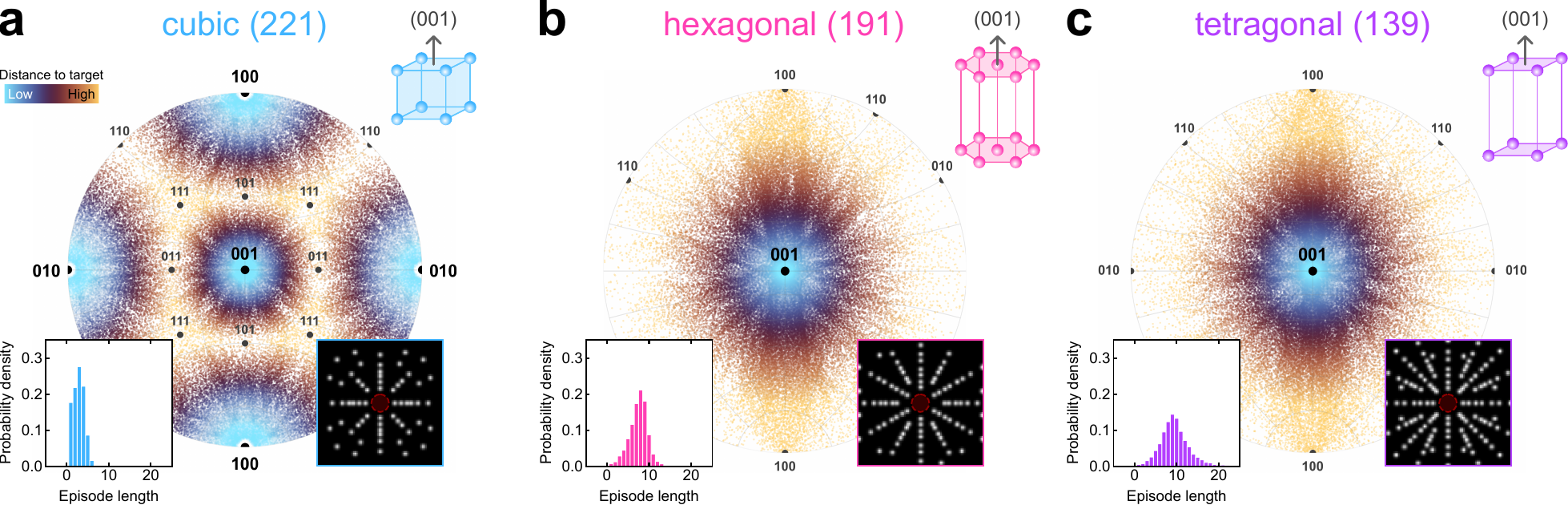}
\caption{\textbf{Evaluation of agent performance}. (a-c) Stereographic projection along the (001) direction for crystal structures (space groups) as indicated, together with 10000 (3000) agent alignment trajectories for cubic (hexagonal and tetragonal) systems. To illustrate the agent's behavior, the step size is here limited to two degrees and a false color scale is used to indicate distance to the closest target. Schematics of the respective crystal structure are given in the top right with high-symmetry (001) axis highlighted. Alignment is achieved when this or equivalent axes are -- within an angular tolerance -- parallel to the x-ray beam. Histograms of number of actions to reach the (001) high-symmetry target (episode length) are shown at the bottom left, obtained by evaluating the agents for 10000 episodes. Insets at the bottom right display examples of the simulated target Laue pattern with angular tolerance (red circles).}
\label{fig:figure_3_performance}
\end{figure*}

\begin{figure*}
\centering
\includegraphics[width=0.95\textwidth]{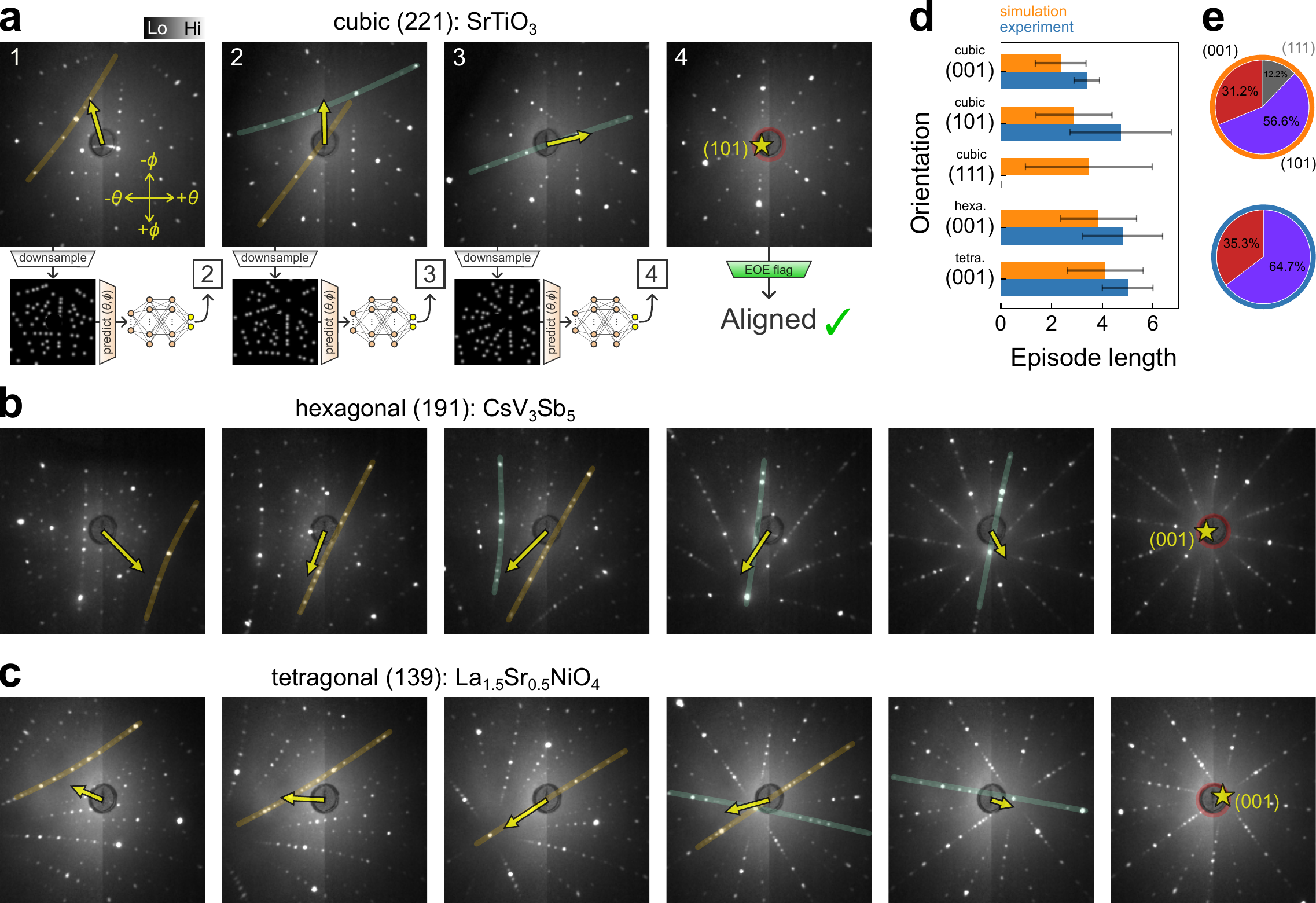}
\caption{\textbf{Agent evaluation on real experimental data}. (a) Representative alignment trajectory obtained on a cubic \STO\ single crystal with space group 221. The top row shows Laue diffraction patterns with examples of high-symmetry lines highlighted in light colors. The patterns are preprocessed and downsampled (lower left images) before being passed as inputs to the agent, which in turn predicts two rotation angles ($\theta,\phi$). Executing those angles using a robotic arm results in a momentum space shift as indicated by the yellow arrows. This sequence repeats until an end-of-episode (EOE) flag is triggered, indicating that the crystal has been aligned along a specific high-symmetry direction (yellow star) within an angular distance of five degrees (red circle). (b,~c) Example trajectories for hexagonal \CVS\ and tetragonal \LSNO\ single crystals with space groups 191 and 139 respectively. In both cases, the agent was trained to target the (001) high-symmetry orientation. (d) Histogram comparing performance of simulation and experiment in terms of average episode length. Error bars represent a single standard deviation. (e) Pie charts of percentage of cubic trajectories targeting specific high-symmetry orientations for simulation (top) and experiment (bottom). For simulation and experiment, statistics in (d,e) are derived from 10000 and 50/25/25 trajectories for cubic/hexagonal/tetragonal crystal structures, respectively.}
\label{fig:figure_4_real_data}
\end{figure*}

\subsection*{Training on simulated data}
\noindent
In Figure~\ref{fig:figure_2_training_curves}, we show training curves for agents, 
trained on mono-atomic cubic, tetragonal, and hexagonal crystal structures.
The defined goal is to align the (001) high-symmetry orientation within a solution tolerance of five degrees. 
In  all cases, the agents quickly converge to 100\% success rate~(\mbox{Figure~\ref{fig:figure_2_training_curves}a}). In the cubic case, the agent requires the least number of actions to reach the target orientation while more steps are required for hexagonal and tetragonal systems~(\mbox{Figure~\ref{fig:figure_2_training_curves}b}). 
This slower convergence 
correlates with lower crystal symmetry 
that in turn reduces the number of high-symmetry targets. In all cases, a fixed set of simulation parameters was used (see Methods section). Additionally, the maximum angular action step size per angle was limited to ten degrees for all agents. 
This was empirically determined to strike a balance between training stability and rapid goal-reaching capability given the used distance-based reward function (see Methods section). The impact of 
different tolerances and action step sizes on training performance is 
shown in \mbox{Supplementary~Figures~S1---S2}.

\noindent
To illustrate the agent behavior, we 
show reciprocal stereographic projections along the (001) direction in \mbox{Figure~\ref{fig:figure_3_performance}}, with high-symmetry orientations superimposed in black. 
Using this representation, we  visualize 
alignment trajectories where the color encodes the distance to the closest high-symmetry target.
In the \mbox{cubic case (Figure~\ref{fig:figure_3_performance}a)}, (100), (010), and (001) directions are equivalent.
Accordingly, the agent can select from multiple high-symmetry targets and typically focuses on the one nearest to its starting position. For \mbox{hexagonal (Figure~\ref{fig:figure_3_performance}b)} and \mbox{tetragonal~(Figure~\ref{fig:figure_3_performance}c)} cases, the agents are trained to find the (001) direction. 
Insets at the top left depict the respective crystal structures with indicated high-symmetry target planes. Insets at the bottom left show histograms of episode length, 
and respective (001) target pattens are shown as insets at the bottom right, superimposed with the solution tolerance (red circles). We observe that the spread of the episode length increases with reduced symmetry (lower space group). The agent trained on the cubic system requires less actions on average, followed by the agent trained on the hexagonal and the tetragonal systems.


\subsection*{Robustness and application to real crystals}
\noindent
We evaluate 
the agent on 
an 
experimental Laue 
setup using poly-atomic single crystals. 
\mbox{Figure~\ref{fig:figure_4_real_data}a} highlights a single alignment trajectory obtained on a \STO\ cubic single crystal. 
We allow the agent to align on any of the cubic high-symmetry orientations, that is (001), (101), or (111). Experimental Laue patterns are 
preprocessed (
binarization and downsampling, see Methods section) 
before being passed through the encoder and actor stages yielding a  rotational action consisting of $(\theta,\phi)$ angles. A new Laue pattern is recorded and the sequence repeated until alignment is achieved. To reduce the simulation-to-real gap, we employ domain randomization~\cite{tobin_domain_2017} (see Methods section).

\noindent
\mbox{Figure~\ref{fig:figure_4_real_data}(b,c)} show examples of alignment trajectories  obtained on hexagonal \CVS\  and tetragonal \LSNO\  single crystals. 
For such 
crystals with reduced symmetry, we find that 
curriculum learning~\cite{narvekar_curriculum_2020} improves the agents performance by making them reach the high-symmetry orientations  within a lower number of steps (see \mbox{Supplementary~Note~3}). In \mbox{Figure~\ref{fig:figure_4_real_data}d}, we compare simulation and experiment in terms of average episode length for individual crystal structures and achieved high-symmetry orientations. While experimental alignments on average take between one and two additional steps, we find an overall excellent agreement between simulation and experiment. 
While the agent occasionally targets the cubic~(111) high-symmetry axis in the simulation, this orientation is not favored in the experiment due to geometric constraints --- see 
Figure~\ref{fig:figure_4_real_data}e.



\subsection*{Discussion}
\noindent
Single crystal alignment relies extensively on human expertise and scattering theory. Supervised learning techniques are, by definition, guided and constrained by human knowledge. As such, the extraction of sample orientation from Laue patterns through these approaches
requires precise 
information on physical parameters that 
govern diffraction patterns~\cite{purushottamrajpurohit_lauenn_2022}. Reinforcement learning, in contrast, enables agents to learn from experience by interacting with complex environments~\cite{mnih_humanlevel_2015,silver_mastering_2017}.
By formulating the process of Laue single crystal alignment as a Markov decision process, we demonstrate that an agent trained through model-free visual reinforcement learning can successfully learn to navigate challenging reciprocal spaces of single crystals from raw pixel observations. In particular, this is achieved without relying on explicit physical models or human-provided action labels.
Similar to humans, the agent autonomously  learns to follow specific high-symmetry lines leading to the desired target orientations. This results in the emergence of crystal-dependent ``highways'', illustrated as regions of increased density in the stereographic projections in \mbox{Figure~\ref{fig:figure_3_performance}}. By training in a simulated environment with mono-atomic crystal structures and applying domain randomization~\cite{tobin_domain_2017} techniques, we demonstrate that the agents achieve excellent results in real-world experimental settings with multi-atomic crystals as highlighted in Figure~\ref{fig:figure_4_real_data}. During training, the episode terminates when the angular distance to the target falls within a defined tolerance. For a real experiment, this end-of-episode trigger is provided by a simple supervised network (see \mbox{Supplementary~Note~4}). 
We also note that the five degrees accuracy can be improved to less than a degree by conventional line-detection algorithms (see \mbox{Supplementary~Note~5}). Such alignment precision is sufficient for the preparation of many different types of scattering experiments~\cite{stock_spin_2008,kuspert_engineering_2024}. Additionally, more effective simulation-to-real transfer~\cite{zhao_simtoreal_2020} can be achieved by applying action-averaging techniques (see \mbox{Supplementary~Note~6}).


\subsection*{Conclusions and outlook}
\noindent
In contrast to supervised approaches that learn mappings from expert-labeled examples, our model-free agent acquires its behavior through interaction and feedback, using only visual observations. This shift from imitation to environment-driven optimization enables the emergence of strategies that are not explicitly encoded by human knowledge. More broadly, this 
paradigm aligns with the view that general intelligence may emerge from autonomous learning through environmental feedback rather than from increasingly sophisticated supervision.\\

\noindent
The autonomous nature of the reinforcement-learning agents introduced here highlights their potential to reduce human efforts in traditionally labor-intensive and expert-driven experimental workflows. An example is the assembly  of single crystal sample mosaics used in neutron spectroscopy experiments. The creation of such mosaics involves precise co-alignment of tens or hundreds of individual single crystals, requiring highly 
repetitive, labor-intensive work~\cite{song_robust_2016}. Consequently, automating this process holds significant potential to reduce both time and labor demands.
  We stress, moreover, that our reinforcement learning methodology is not restricted to laboratory Laue diffraction --- it can be generalized to other scattering techniques.
 Electron and synchrotron x-ray diffraction experiments 
very often begin with alignments of samples, which 
is time consuming even for experienced operators. Automation of these processes will therefore substantially increase efficiency, a fundamental factor in the time-constrained environment of large-scale facilities.



\subsection*{Methods}
\vspace{-1cm}
\noindent
\subsubsection*{Software and hardware}
\vspace{-0.25cm}
\noindent
All code was written in Python~3.9 with PyTorch~\cite{paszke_pytorch_2019} as the machine learning framework. Training of the Laue agents was carried out using an Nvidia~A100, 4~cores, and 64~GB~of~memory.

\vspace{-0.25cm}

\subsubsection*{Reinforcement learning}
\vspace{-0.25cm}
\noindent
In a Markov decision process~(MDP)~\cite{bellman_markovian_1957} --- modeling the interaction between an agent and an environment ---
an agent observes a representation of an environment state $S_t \in \mathcal{S}$. Based on this observation, the agent selects an action $A_t \in \mathcal{A}$. As a consequence of the chosen action, the agent receives a reward signal at the next time step $R_{t+1}$ while transitioning into a new state $S_{t+1}$. We denote $\mathcal{S}$ as the state space and $\mathcal{A}$ as the action space. To improve sample efficiency, we adapted DrM~\cite{xu_drm_2023} as the reinforcement-learning~(RL) algorithm with pixel observations.
The process of aligning a single crystal along a high-symmetry direction using Laue diffraction and independent rotation angles can be formulated as a fully observable MDP with a static environment. That is, the state (two-dimensional Laue pattern) does not change unless the agent performs an action. As such, we neither use frame stacking nor action repeat. 
In Table~\ref{tab:rl_parameters} we list the used RL hyperparameters. For the  encoder, we follow  the original convolutional-neural-network architecture proposed in~\cite{mnih_playing_2013} with slight modifications -- see Table~\ref{tab:encoder_structure}. After passing the encoder stage, the flattened output of the last convolutional layer is first compressed to a feature vector with 50 neurons, followed by a layer normalization and a Tanh activation function. In \mbox{Supplementary~Note~7} we compare the performance of different RL algorithms. 

\begin{table}[h!]
\centering
\caption{DrM~\cite{xu_drm_2023} hyperparameters for the RL training.}
\begin{tabular}{l c}
\hline
\textbf{Parameter} & \textbf{Setting}\\
\hline
Observation shape & 84 x 84\\
Frame stack & 1\\
Action repeat & 1\\
Augmentation padding & 4\\
Discount rate $\gamma$ &  0.99\\
Seed frames & 2000\\
Exploration steps & 1000\\
Training steps (cubic) & 200000\\
Training steps (tetra., hexa.) & 300000\\
Replay buffer capacity & same as training steps\\
Evaluation episodes & 100\\
$n$-step returns & 3\\
Mini-batch size & 256\\
Gradient descent optimizer & Adam\\
Learning rate & 0.0001\\
Agent update frequency & 1\\
Soft update rate & 0.01\\
Feature dimension & 50\\
Hidden dimension & 1024\\
Perturb rate & 1\\
Dormant threshold & 0.025\\
Target dormant ratio & 0.2\\
Minimum perturb factor & 0.2\\
Maximum perturb factor & 1.0\\
Dormant perturb interval & 20000\\
Dormant temperature & 10\\
Target exploitation parameter & 0.6\\
Exploitation expectile & 0.9\\
Std. dev. type & awake\\
Std. dev. schedule & linear(1.0, 0.1, 100000)\\
Std. dev. clip & 0.3\\
\hline
\end{tabular}
\label{tab:rl_parameters}
\end{table}

\begin{table}[h!]
\centering
\caption{Encoder structure used for extracting features from the Laue patterns. Order of convolutional layer parameters are (filters, kernel size, stride, padding). After each convolutional layer we apply a ReLU activation function to introduce non-linearity.}
\begin{tabular}{l c c}
\hline
\textbf{Layer} & \textbf{Input shape} & \textbf{Output shape}\\
\hline
Conv2D(32, 3, 2, 1) & (1, 84, 84) & (32, 42, 42)\\
Conv2D(32, 3, 2, 1) & (32, 42, 42) & (32, 21, 21)\\
Conv2D(32, 3, 1, 1) & (32, 21, 21) & (32, 21, 21)\\
Conv2D(32, 3, 1, 1) & (32, 21, 21) & (32, 21, 21)\\
\hline
\end{tabular}
\label{tab:encoder_structure}
\end{table}

\begin{table*}
\centering
\begin{minipage}{0.65\textwidth}
\caption{Simulation parameters. The second column indicates values for a fixed setting while the third column displays value ranges for domain randomization.}
\setlength{\tabcolsep}{10pt}
\begin{tabular}{l c c}
\hline
\textbf{Parameter} & \textbf{Fixed} & \textbf{Domain randomization}\\
\hline
Detector-film distance (cm) & 5 & 4 -- 6\\
Number of spots (cubic) & 60 & 30 -- 60\\
Number of spots (hexagonal) & 90 & 40 -- 90\\
Number of spots (tetragonal) & 120 & 60 -- 120\\
Cubic lattice constants (\AA) & 9 ($a,b,c$) & 3 -- 15 ($a,b,c$)\\
Hexa. lattice constants (\AA) & 4 ($a,b$), 8 ($c$) & 4 -- 7 ($a,b$), 8 -- 11 ($c$)\\
Tetra. lattice constants (\AA) & 3 ($a,b$), 12 ($c$) & 3 -- 6 ($a,b$), 12 -- 15 ($c$)\\
Spot shift $\sigma$ & 0 & 1.0\\
Spot removal fraction & 0 & 0.25\\
Spot placement fraction & 0 & 0.0 -- 0.1\\
\hline
\end{tabular}
\label{tab:agent_parameters}
\end{minipage}
\end{table*}

\vspace{-0.25cm}

\subsubsection*{Reward function design}
\vspace{-0.25cm}
\noindent
The learning effectiveness of a reinforcement-learning agent strongly depends on the chosen reward function. A reward acts as a guiding signal for the agent's learning and should therefore be as informative as possible. As such, we define a reward function $R_t$ for each time step $t>0$ based on the angular distance $d_t$ to the closest high-symmetry target $R_t = \frac{100(d_{t-1} - d_t)}{d_0 \sqrt{t}}$.
If the agent is able to reach the high-symmetry target within 50 steps and an angular solution tolerance of five degrees, 
additional $+100$ is rewarded.
We find that this reward function results in faster and more reliable learning compared to sparse rewards, where the agent is given a constant step penalty (see \mbox{Supplementary~Note~8}).

\vspace{-0.25cm}

\subsubsection*{Laue environment}
\vspace{-0.25cm}
\noindent
Laue patterns are calculated using information about the crystal, such as space group and lattice constants $(a,b,c,\alpha,\beta,\gamma)$, and detector geometry including sample-detector distance, detector size, and energy of the x-ray beam. 
The original Python code (Ref.~\cite{bailey_advanced_2016}) has been extended to simulate arbitrary mono-atomic crystal structures
and its performance has been optimized to facilitate the RL training. A custom Laue environment was implemented using the DeepMind control suite~\cite{muldal_dm_env_2019,tunyasuvunakool_dm_control_2020}. 
The environment emulates an alignment system where the orientation of the crystal is defined by three angles $(\theta,\chi,\phi)$. Changing $\theta$ (yaw) or $\phi$ (pitch) will result in a lateral shift of the Laue pattern on the detector while $\chi$ (roll) corresponds to a rotation of the Laue pattern around the x-ray beam axis. For simplicity, we disregard $\chi$ during the alignment stage but note that $\chi$ can be easily corrected once the $(\theta,\phi)$ angles have been fixed.
At each time step $t$, the agent predicts $(\theta_t,\phi_t)$ action values from the action space $\mathcal{A} \in [-1, 1]$, which are scaled to the maximum angular action range $\mathcal{A}_{s} \in [-10,10]$ degrees. This decision, combined with the chosen reward function (see Methods section \textit{Reward function design}), is made to ensure that the maximum number of actions needed for alignment remains as low as possible. We also find that having a larger maximum angular action step size results in both unstable training and worse performance (see \mbox{Supplementary~Figure~S2}). We attribute this to the fact that consecutive observations, separated by larger angular distances, show reduced visual correlation and therefore provide the agent with less information about the state-action relationship.\\

\noindent
We then define the crystal orientation at each time step~$t$ using a 
rotation matrix $\Omega_t$ as

\vspace*{-0.4cm}

\begin{align*}
\Omega_t &= \Omega_0 R_y(\theta_t) R_z(\phi_t) \Omega_{0}^{T}\\
\Omega_0 &= R_y(\theta_0) R_x(\chi_0) R_z(\phi_0)
\end{align*}

\noindent
where $R_x$, $R_y$, and $R_z$ correspond to rotation matrices around the $x$, $y$, and $z$ axis respectively. $\Omega_0$ is re-created in the beginning of each  episode by centering on a high-symmetry target and sampling three initial angles $(\theta_0,\chi_0,\phi_0)$ from a uniform distribution of the maximum angular initial range \mbox{$\mathcal{I} (\theta_0,\phi_0) \in [-90,90] \, \mathrm{degrees}$} and \mbox{$\mathcal{I}(\chi_0) \in \left[ -180, \, 180 \right] \, \mathrm{degrees}$}. 
To enable the agents to better explore we allow a total angular action range $\mathcal{A}_{\mathrm{max}} (\theta,\phi) \in [-120,120]$ degrees. The second column of Table~\ref{tab:agent_parameters} indicates the fixed simulation parameters used for the agents highlighted in Figure~\ref{fig:figure_2_training_curves} and Figure~\ref{fig:figure_3_performance}. 
When training the agents for the experimental measurements, both $\mathcal{I} (\theta_0, \phi_0)$ and $\mathcal{A}_{\mathrm{max}} (\theta, \phi)$ ranges were halved. This was done in response to the various geometrical constraints of the used Laue setup such as the limited chamber size and the maximum reach of the robot arm joints.
Given the 
matrix $\Omega_t$, the positions of the Laue spots on the detector can be calculated as described by Preuss et al.~\cite{preuss_laue_1974}. We then select a random number of most intense Laue spots and render a binary image of size 1284~x~1284 pixels where each Laue spot has a radius of 10 pixels. This high-resolution image is then downsampled to 84~x~84 pixels using a box filter before being passed to the agent. Finally, the images are blurred with a Gaussian filter ($\sigma=1$) and the central x-ray camera pin-hole region is removed (radius 10\% of image size).

\subsubsection*{Domain randomization}
\vspace{-0.25cm}
\noindent
Laue agents are trained on individual crystal structures with specific space groups. During training, we employ domain randomization~\cite{tobin_domain_2017} to create agents that can generalize well on a wide range of experimental settings. This includes varying lattice parameters, sample-detector distance, and number of Laue spots. We furthermore randomly offset Laue spots from their theoretical positions, randomly remove a fraction of the Laue spots, and randomly place a number of spots to emulate 
spurions. For the cubic model, we use additional domain randomization by random space group variation during training (221, 225, 229, and mix of 225 and 229).  
These data operations are made to make the agents robust against real experimental conditions. 
The third column of Table~\ref{tab:agent_parameters} shows the range of parameters used for domain randomization.
More details about the generalization capabilities of the agent for different cubic space groups can be found in \mbox{Supplementary~Note~9}.

\vspace{-0.25cm}

\subsubsection*{Experimental Laue patterns}
\vspace{-0.25cm}
\noindent
We used a commercial x-ray Laue instrument from Photonic Science with a Meca500 six-axis robot arm from Mecademic (see Supplementary~Figure~S12). We typically operated with a crystal-detector distance of roughly 4~cm and used an x-ray tube voltage of 50 kV. For testing of the network, we used single crystals of SrTiO$_3$ (space group 221), CsV$_3$Sb$_5$ (space group 191), and La$_{1.5}$Sr$_{0.5}$NiO$_4$ (space group 139). Exposure times per Laue pattern were 60, 120, and 60 s respectively. The recorded diffraction patterns are of size 1940~x~1284 pixels and were saved in the TIF image format. Laue patterns used by the trained agent, were subject to the following preprocessing steps. We first apply a median filter with one pixel width, followed by Gaussian smoothing with a standard deviation of 0.5 to remove dead pixels. Intensities are then clipped to the upper 99\% percentile to reduce strong outliers. Finally, we identify the Laue spots positions from the processed Laue patterns using the Laplacian of Gaussian method. The determined spot positions are then used to render the agent input image following the same procedure as for the simulation (see Methods section \textit{Laue environment}).

\vspace{0.25cm}

\noindent
\textit{Data Availability:}
The experimental data used in this work  is available at \mbox{\href{https://doi.org/10.5281/zenodo.19485314}{10.5281/zenodo.19485314}}.\\

\noindent
\textit{Code Availability:}
The code used for the Laue simulation and reinforcement-learning training  is available at \mbox{\href{https://github.com/joppli/laue_rl}{github.com/joppli/laue\_rl}}.\\

\noindent
\textit{Acknowledgements:} 
J.O. acknowledges support from a Candoc grant of the University of Zurich (Grant No. FK-22-095). D.P. acknowledges the EPSRC, UK  and the Oxford-ShanghaiTech collaboration project for financial support. Q.W. is supported by the Research Grants Council of Hong Kong (ECS No. 24306223), and the Guangdong Provincial Quantum Science Strategic Initiative (GDZX2401012).\\

\noindent
\textit{Author Contributions:} J.~O. conducted the entire project, from conceptualizing the methodology to executing  the work. J.~O. implemented, tested, and executed the reinforcement learning algorithm. Simulation software of Laue patterns was upgraded by J.~O. and M.~S.. All experimental Laue recordings were carried out by J.~O. with assistance from A.~R., I.~B. and L.~M.. \LSNO\ and \CVS\ crystals were provided by D.~P., Q.~W. and J.~Z.. P.G.F. characterized the nickelate crystals. The manuscript was written by J.~O. and J.~C. with input from all co-authors.\\

\noindent
\textit{Competing Interests:}
The authors declare no competing interests.


\bibliography{Ref-LQMR-Jens}

\end{document}


\beginsupplement
%
%

\title{\Large \textbf{Supplementary Information}\\ \vspace{0.25cm}
\Large Autonomous Diffractometry Enabled by Visual Reinforcement Learning}
      
\author{J.~Oppliger \textit{et al.}}
\date{}

\maketitle

\section{Introduction to Laue x-ray diffraction}

Significant technological advancements are oftentimes preceded by breakthroughs in materials research. As such, the development of floating zone techniques~\cite{pfann_principles_1952} enabled the production of semicondutor-grade silicon, leading to the realization of the first transistors --- essential to most of the electronic devices today. More examples include the discovery of lithium cobalt oxides for the production of batteries with high energy densities~\cite{mizushima_lixcoo2_1980} --- crucial for electric vehicles and energy storage in general --- and the discovery of novel superconducting materials~\cite{bednorz_possible_1986,berlincourt_emergence_1987}, giving rise to medical magnetic resonance imaging 
and dissipationless transmission of electricity. 
Progress in modern materials science is directly governed by fundamental condensed matter research, which provides both theoretical and experimental foundations for understanding and designing materials with novel structural and electronic properties. Nowadays, a major part of condensed matter research focuses on studying physics underlying quantum materials. Experimentally, such materials are often grown in the form of high-quality single crystals and studied by diffraction, scattering, or spectroscopy techniques at large-scale synchrotron facilities involving probing particles such as photons, electrons and neutrons~\cite{kuspert_engineering_2024, simeth_microscopic_2023}. This allows for direct investigation of a vast range of structural and strongly correlated electronic phases, many of which take place in specific crystallographic lattice planes~\cite{tranquada_evidence_1995,kenzelmann_coupled_2008,vibhakar_spontaneous_2024}. Because of that, precise knowledge and control of the crystal orientation is essential for performing state-of-the-art experimental research. One critical step during the sample preparation step is therefore the alignment of the sample, where the single crystal has to be oriented along a crystallographic (high-symmetry) direction.\\

\noindent
The Laue x-ray diffraction method is commonly used to guide the alignment process of a single crystal by means of determining crystallographic orientations. In this method, a polychromatic x-ray beam scatters elastically the crystal, and the resulting diffraction intensities are recorded on a two-dimensional detector. While there exist different experimental Laue geometries, the here presented work focuses on the so-called back-reflection geometry. In accordance with Bragg's law and the crystal symmetry, the recorded diffraction pattern comprises discrete reflections (Laue spots) corresponding to lattice planes characterized by their Miller indices $(hkl)$. The alignment of single crystals is oftentimes performed in a labor-intensive manual fashion, but can be assisted by a range of well-established software analysis tools~\cite{ouladdiaf_orientexpress_2006,schumann_cologne_2011,micha_lauetools_2009,huang_upgraded_2023}. Traditionally, such programs deploy a sequence of spot fitting and indexation (assignment of Miller indices to spots). The classical indexation process operates by means of identifying pairs of experimental peaks using prior specification of physical parameters such as lattice constants~\cite{riquet_depouillement_1979}. This is followed by the calculation of several theoretical candidate Laue patterns, which are compared to the experimental  data and yield a so-called orientation matrix~\cite{busing_angle_1967}. The latter allows to perform the alignment of the crystal along a particular crystallographic plane by  establishing a transformation between Miller indices and rotation angles. Recently, a machine-learning method has been proposed to speed-up traditional spot indexation and facilitating the analysis of Laue microdiffraction patterns by means of supervised training of a fully-connected neural network~\cite{purushottamrajpurohit_lauenn_2022}. A major bottleneck of all of the aforementioned methods is their strong dependence on the accurate knowledge of system parameters such as lattice constants, unit cell composition and detector geometry. 
Moreover, for traditional approaches~\cite{ouladdiaf_orientexpress_2006,schumann_cologne_2011,micha_lauetools_2009,huang_upgraded_2023}, human input is typically required to provide initial estimates of the Laue spot locations. These dependencies inherently limit the degree of automation achievable in the crystal alignment process.

\clearpage

\section{Influence of angular solution tolerance and action step size}

\begin{figure}[h!]
\centering
\includegraphics[width=0.8\textwidth]{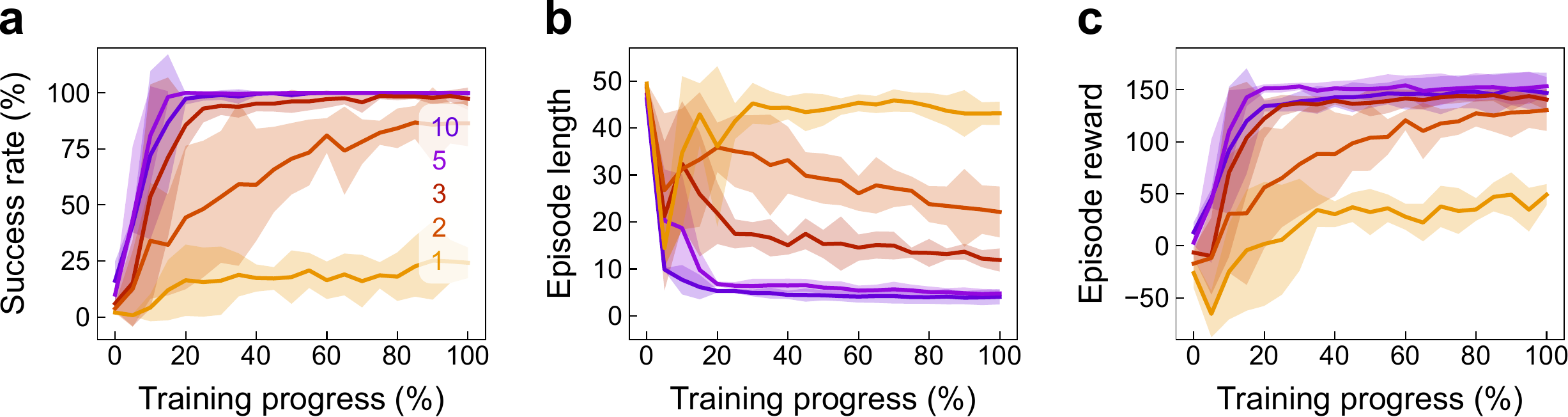}
\caption{\textbf{Impact of different angular solution tolerances.} (a-c) Training curves for a cubic system showing episode reward, episode length, and success rate as a function of percentage training cycles for different angular solution tolerances in degrees as indicated. 100\% training progress corresponds to 100000 training steps. Solid lines represent the mean, while shaded areas indicate the 95\% confidence interval calculated from five training runs with different random seeds.}
\label{fig:figure_SI_varying_maxdist}
\end{figure}

\begin{figure}[h!]
\centering
\includegraphics[width=0.8\textwidth]{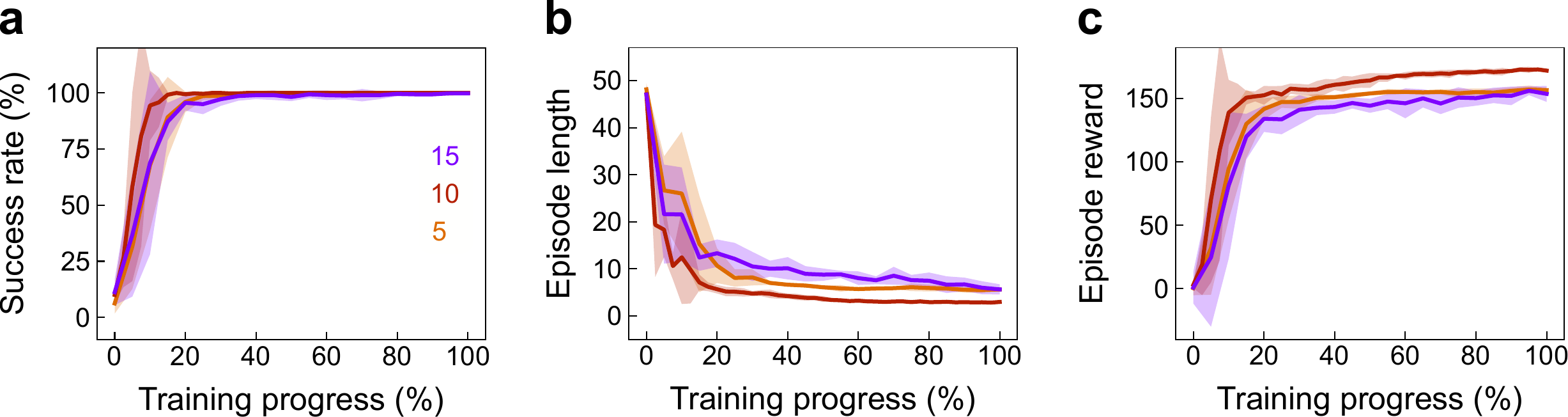}
\caption{\textbf{Impact of different maximum angular action step sizes}. (a-c) Training curves for a cubic system showing episode reward, episode length, and success rate as a function of percentage training cycles for different maximum angular action step sizes in degrees as indicated. 100\% training progress corresponds to 100000 training steps. Solid lines represent the mean, while shaded areas indicate the 95\% confidence interval calculated from five training runs with different random seeds.}
\label{fig:figure_SI_varying_max_action_change}
\end{figure}

\clearpage

\section{Curriculum learning}

\begin{figure}[h!]
\centering
\includegraphics[width=0.95\textwidth]{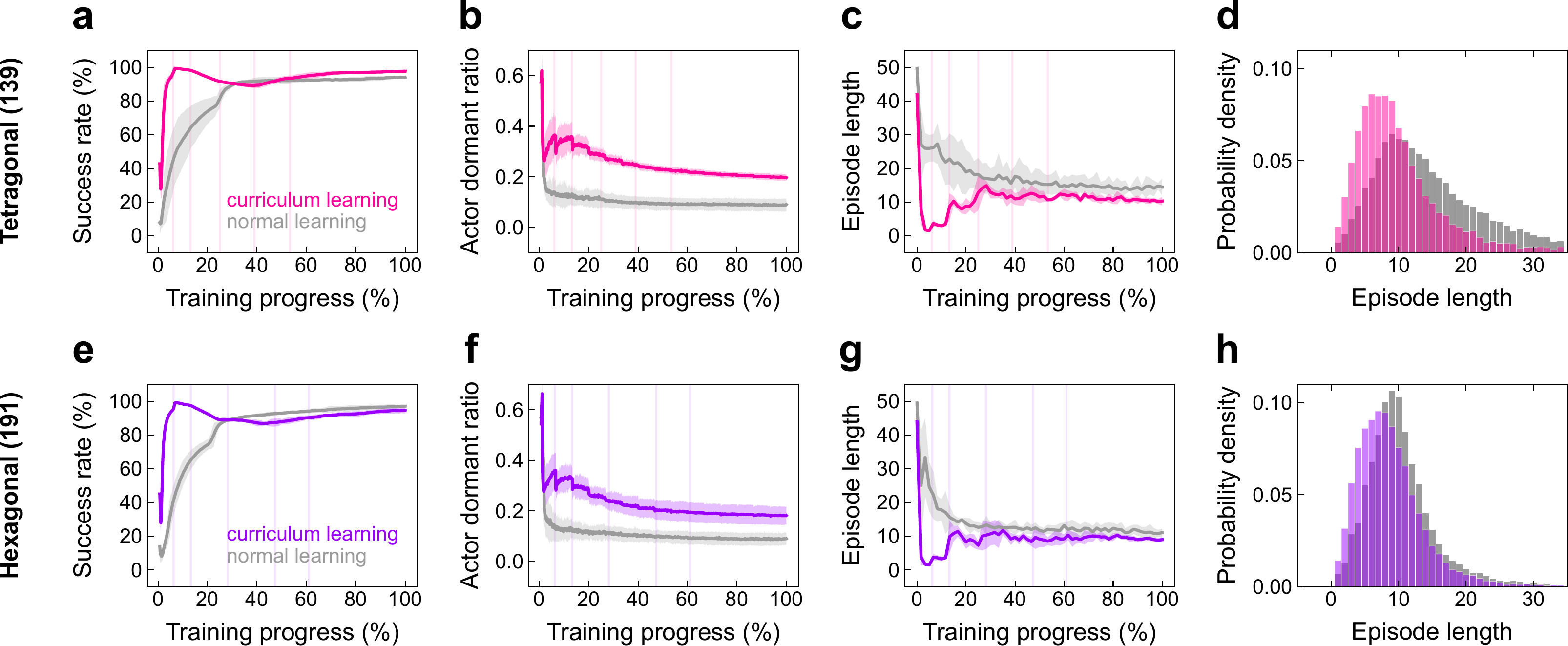}
\caption{\textbf{Effect of curriculum learning on training performance}. Performance comparison of agents trained on tetragonal (a-d) and hexagonal (e-h) crystal structures with domain randomization. Solid lines in (a-c) and (e-g) represent the mean, while shaded areas indicate the 95\% confidence interval calculated from five training runs with different random seeds. The faint vertical lines mark the positions at which the curriculum learning step has occurred. Gray lines represent the non-curriculum learning baseline. (d,h) Histograms of episode lengths for the respective agents evaluated for 10000 episodes.}
\label{fig:figure_SI_curriculum_learning}
\end{figure}

For training of the baseline agents, the angular initial range is fixed for the entire training duration as the maximum angular initial range \mbox{$\mathcal{I} (\theta_0, \phi_0) \in \left[ -90, \, 90 \right] \, \mathrm{degrees}$} and \mbox{$\mathcal{I}(\chi_0) \in \left[ -180, \, 180 \right] \, \mathrm{degrees}$}. A curriculum learning (CL) scheme~\cite{narvekar_curriculum_2020} starts with a smaller initial range, \mbox{$\mathcal{I}_{0} (\theta_0,\phi_0) \in [-15,15]$} degrees, which is then symmetrically increased by 15 degrees when the moving average of the success rate  over 3000 training episodes exceeds 90\%. This cycle repeats until the final initial angular range, \mbox{$\mathcal{I}_{5} (\theta_0,\phi_0) \in [-90,90]$} degrees, is reached. In Figure~\ref{fig:figure_SI_curriculum_learning}, we show training curves for both tetragonal and hexagonal crystal structures with domain randomization as explained in the Methods section of the main text. Figure~\ref{fig:figure_SI_curriculum_learning}(a,e) and Figure~\ref{fig:figure_SI_curriculum_learning}(b,f) show the training success rate and actor dormant ratio~\cite{sokar_dormant_2023}, the latter being a metric of the fraction of dormant neurons within the actor network. Generally, a low dormant ratio indicates efficient learning. Figure~\ref{fig:figure_SI_curriculum_learning}(c,g) show the average episode length over 100 evaluation episodes, highlighting how a CL schedule helps in reducing the average episode length of the agents compared to normal learning for crystal structures of lower symmetry (lower space group number). Finally, histograms of episode lengths in Figure~\ref{fig:figure_SI_curriculum_learning}(d,h) demonstrate that a consistent reduction of the average episode length can be achieved using CL in this way.

\clearpage

\section{Supervised high-symmetry identification}


During the reinforcement-learning~(RL) training stage, the agent receives an end-of-episode signal after reaching a desired high-symmetry orientation. In a real experimental setting, such a signal is usually provided by a human expert. This poses no significant problem when the goal is to align single crystals with human oversight. Nevertheless, there  exist situations where a higher degree of automation is desired. For example, when many individual single crystals have to be co-aligned to create a single crystal mosaic for inelastic neutron scattering experiments~\cite{stock_spin_2008,steffens_spin_2019,song_robust_2016,lane_twodimensional_2021,stock_ising_2018,stone_extended_2003,huvonen_excitations_2012,shen_evidence_2016,ma_disorderinduced_2021,simeth_microscopic_2023}. 
In such scenarios, the agent should not have to rely on human input to stop the alignment process.\\ 

\begin{figure}[h!]
\centering
\includegraphics[width=0.95\textwidth]{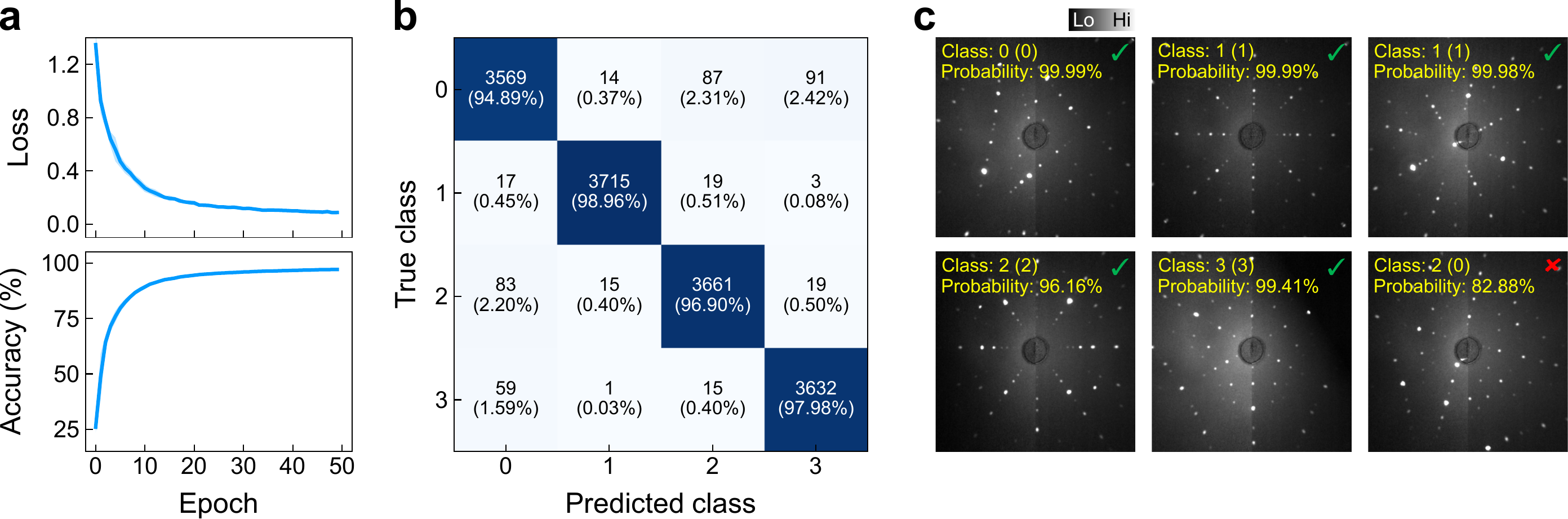}
\caption{\textbf{Supervised high-symmetry identification}. (a)~Validation categorical cross entropy loss (top) and validation accuracy (bottom) as a function of training epochs of a supervised CNN trained on the cubic crystal structure with space groups 221, 225, and 229.
Solid lines represent the mean, while the (barely visible) shaded area the 95\% confidence interval obtained using 5-fold cross validation.
(b)~Multiclass confusion matrix, evaluated on a separate simulated test set. Class 1, 2, and 3 refer to (001), (101), and (111) high-symmetry axes respectively, within a solution tolerance of five degrees. Class 0 encompasses any other (non high-symmetry) orientation. (c)~Trained network applied to experimental Laue patterns recorded on a SrTiO$_3$ single crystal with predicted (true) class labels and probabilities as indicated. A green tick (red cross) mark in the top right indicates a successful (false) prediction.
}
\label{fig:figure_SI_supervised_prediction}
\end{figure}

\noindent
For this purpose, we created an automated decision-making process that provides the agent with an end-of-episode signal as soon as a desired high-symmetry orientation is reached. By considering the task of high-symmetry identification as a multiclass classification problem, we train a small convolutional neural network (CNN) on a labeled simulated Laue diffraction dataset. The Laue patterns are generated following the same procedure as during the training stage of the RL agent with domain randomization~\cite{tobin_domain_2017}. The architecture of the convolutional layers is identical to the structure of the agent encoder stage.
As in the RL process, the flattened output of the last convolutional layer is compressed to a 50-neuron feature vector, followed by a layer normalization and a Tanh activation function. Afterwards, the output is passed through two fully-connected layers with 64 neurons each before reaching the final classification layer, formed by four neurons. The number of training, validation, and testing samples was 48000, 12000, and 15000. An Adam optimizer with a learning rate of $5 \cdot 10^{-5}$ and a batch size of eight was used.
In Figure~\ref{fig:figure_SI_supervised_prediction}, we show results for identifying different high-symmetry orientations for the cubic crystal structure. Before being passed to the trained CNN, the Laue patterns are preprocessed and downsampled in the same fashion as the inputs of the reinforcement-learning agent.
We generally observe a high model accuracy in predicting the correct classes for both simulated and experimental patterns.
Sometimes, the model might misclassify a state as shown in the bottom right example of Figure~\ref{fig:figure_SI_supervised_prediction}(c) where it predicted a (101)-style high-symmetry orientation (class 2), which is in fact not a high-symmetry orientation (class 0). However, such cases are rare and usually governed by low prediction probabilities, which enables simple probability thresholding techniques, e.g. only accept a prediction if the probability is greater than 95\%.

\clearpage

\section{Procedure for  high-accuracy crystal alignment}

\begin{figure}[h!]
\centering
\includegraphics[width=0.95\textwidth]{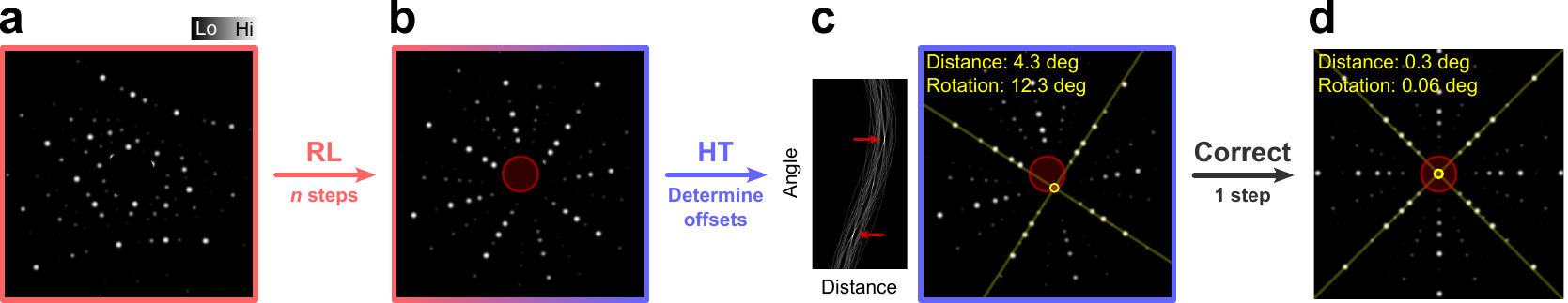}
\caption{\textbf{Procedure for high-accuracy crystal alignment}. (a-b) A reinforcement-learning~(RL) agent, described in the main text, is  aligning a single crystal along a given high-symmetry orientation within $n \geq 1$ steps and a  distance tolerance of five degrees (indicated by the red circle). (c) Once a high-symmetry orientation is found, the Hough transform~(HT) is used to calculate the angular offsets by selecting the (angle,~distance) parameters with highest probabilities in the Hough parameter space, indicated by red arrows in the inset. The high-symmetry lines detected by the HT are highlighted in yellow on top of the  Laue pattern. (d) The angular offsets determined in (c) are then used to achieve a high-accuracy alignment after a single step.}
\label{fig:figure_SI_fine_alignment}
\end{figure}

After the alignment procedure using the reinforcement-learning~(RL) agent, the crystal is oriented along a high-symmetry direction within an angular solution tolerance of five degrees. While five degrees is generally sufficient to provide a human experimentalist with a rough guideline,
many scattering experiments require tolerances in the range of one degree or lower. The chosen tolerance of five degrees was found to strike a good balance between stable training and fast alignment when using a step size of ten degrees (see Figures~\ref{fig:figure_SI_varying_maxdist} and \ref{fig:figure_SI_varying_max_action_change}). A smaller  tolerance was found to lead to higher likelihood of oscillatory behavior around the target state. However, RL is actually not required to perform the final high-accuracy alignment once the high-symmetry state is situated close to the detector center. In this case, the high-symmetry crystallographic directions are significantly less parabolic, which enables the application of efficient conventional line detection algorithms such as the Hough transform~\cite{hough_method_1962,duda_use_1972}. 
In Figure~\ref{fig:figure_SI_fine_alignment}, we show the sequence of initial alignment using an RL agent, followed by the application of the Hough transform to identify the remaining angular offsets. The knowledge of these offsets allows to perform a final high-accuracy alignment using a single additional step. For further reducing the remaining angular offsets, the Hough transform can be applied iteratively. 

\section{Action averaging}

\begin{figure}[h!]
\centering
\includegraphics[width=0.95\textwidth]{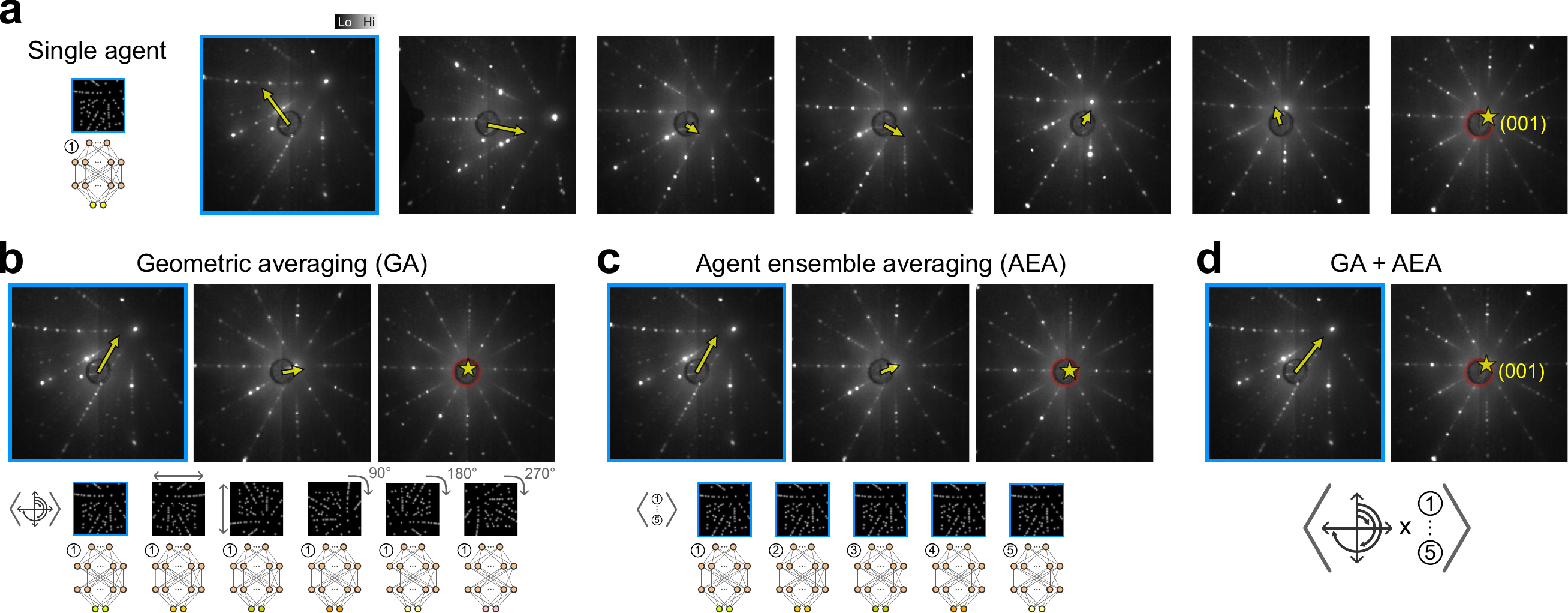}
\caption{\textbf{Approaches for high-confidence action prediction}. (a) Experimental Laue alignment trajectory on a hexagonal CsV$_3$Sb$_5$ single crystal as predicted by a single agent. The yellow arrows indicate the predicted action while the yellow star highlights the target state within the solution tolerance (red circle). (b,c) Trajectory achieved with the same starting pattern (blue border) as in (a) where the action taken for the current image is calculated as the average over predictions of six unique transformations or over predictions of five agents trained with different random seeds. 
(d) Trajectory achieved using a combination of 
approaches in (b) and (c).}
\label{fig:figure_SI_action_averaging}
\end{figure}

Trained agents generally achieve consistent high performance in reaching target orientations on real single crystals. Yet, we find that they sometimes display erratic behavior. This refers to situations where the agents are oscillating around a particular state or are moving towards the target orientation along an indirect path --- see Figure~\ref{fig:figure_SI_action_averaging}a. We propose three methods to increase and judge the confidence of an action prediction. First, we note that the agent encoder is comprised of convolutional layers where the filters are inherently translation equivariant. Given the random sampling of initial crystal orientation angles during training, the filters should furthermore offer approximate rotation and mirror invariance. 
Taking inspiration from test-time augmentation (TTA)~\cite{ayhan_testtime_2022,shanmugam_better_2021} --- usually deployed to increase accuracy in classification tasks --- we introduce geometric averaging (GA). Here we apply a single trained agent to a set of transformed versions of the original image as shown in Figure~\ref{fig:figure_SI_action_averaging}b. The final action is then calculated as a simple average. A second approach, denoted as agent ensemble averaging (AEA), uses an ensemble of five agents trained with different random seeds, whose action predictions for the original input are averaged as illustrated in Figure~\ref{fig:figure_SI_action_averaging}c. Both GA and AEA methods reduce the variance of a single agent prediction  and generally offer a more efficient alignment process, effectively reducing the number of steps required. Finally, we highlight the achieved trajectory obtained by combining  GA and AEA averaging methods in Figure~\ref{fig:figure_SI_action_averaging}d. We furthermore expect that such averaging techniques could also serve as confidence measures. By computing the similarity between individual action predictions, a confidence metric could be defined that reflects the reliability of the selected action.\\

\begin{figure}[h!]
\centering
\includegraphics[width=0.95\textwidth]{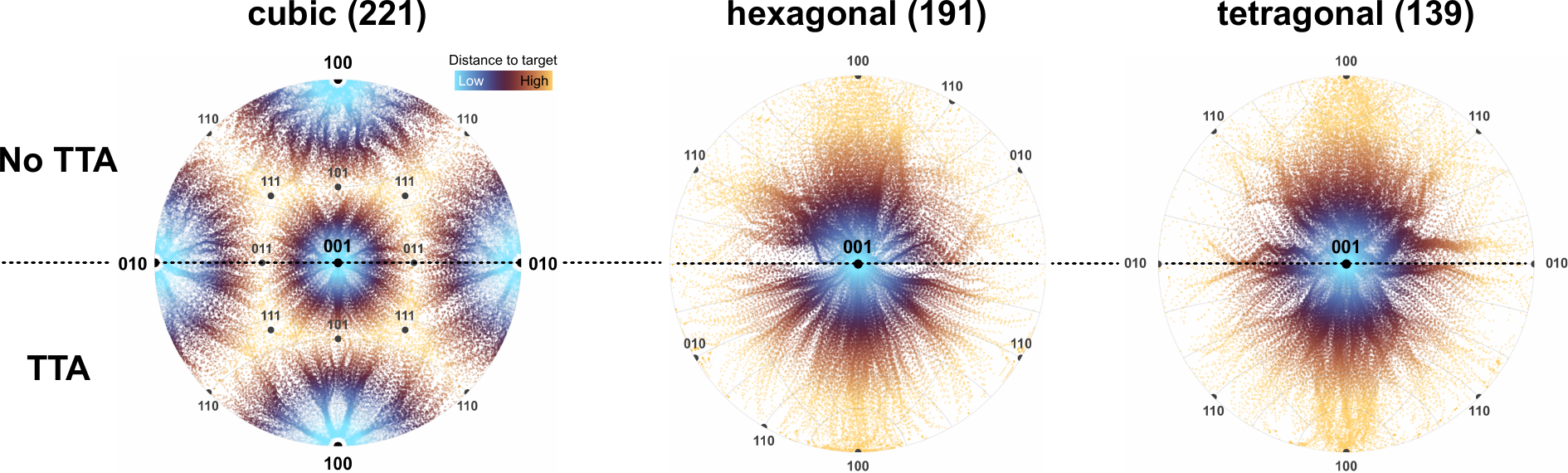}
\caption{\textbf{Single agent trajectories with and without test-time data augmentation (TTA).} Top panels (above dashed line) display trajectories of a single agent trained to operate in cubic, hexagonal, or tetragonal reciprocal space with space groups as indicated and evaluated for a fixed $\chi_0$ orientation. Bottom panels (below dashed line) display the same agents but with decision trajectories obtained through an average of mirror and rotational augmentation of the Laue patterns. }
\label{fig:figure_SI_geometric_average}
\end{figure}

\noindent
Overall, the effectiveness of the approaches discussed above can be attributed to a grounding effect of averaging the agent’s predicted actions. In Figure~\ref{fig:figure_SI_geometric_average}, we compare the trajectories of a single trained agent with and without TTA. In contrast to Figure~3 in the main text, where the initial $\chi_0$ orientation was randomly chosen, we here fix the $\chi_0$ orientation for each episode at $\chi_0 = 0$. Without TTA, this results in more pronounced ``highways'' in reciprocal space, which show a clear asymmetry, highlighting that the agent has learned to follow different high-symmetry lines depending on its direction (top part of Figure~\ref{fig:figure_SI_geometric_average}). By performing  TTA in the form of GA, we observe the emergence of highly symmetric ``highways'' around the target orientations (bottom part of Figure~\ref{fig:figure_SI_geometric_average}). This indicates that more uniform and targeted trajectories can be achieved using action averaging. This is especially useful when the trained agents are deployed in a real experimental setup where possible differences to the training phase are more likely to lead to single agent predictions of higher variance.

\clearpage

\section{Reinforcement-learning algorithms}

\begin{figure}[h!]
\centering
\includegraphics[width=0.8\textwidth]{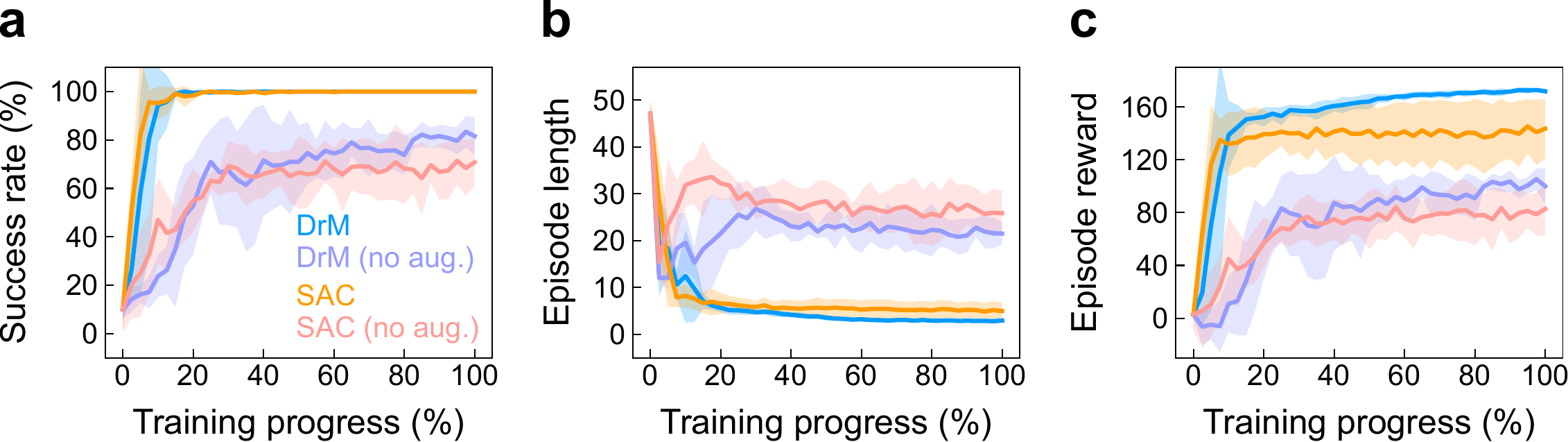}
\caption{\textbf{Comparison of different reinforcement-learning algorithms}. (a-c) Training curves for a cubic system as a function of percentage training cycles for different algorithms with and without data augmentation. Solid lines represent the mean, while shaded areas indicate the 95\% confidence interval calculated from five training runs with different random seeds.}
\label{fig:figure_SI_rl_algos}
\end{figure}

In this work, we used dormant ratio minimization (DrM)~\cite{xu_drm_2023}, one of the state-of-the-art model-free \mbox{reinforcement-learning~(RL)} algorithms for learning directly from pixels. DrM builds on the success of soft-actor-critic (SAC) algorithms~\cite{haarnoja_soft_2018,yarats_improving_2021} combined with improved data augmentation techniques~\cite{laskin_reinforcement_2020,yarats_image_2020,yarats_mastering_2021} to alleviate the problem of spatially inconsistent gradients in the convolutional encoder that typically arise in temporal difference learning from pixels~\cite{cetin_stabilizing_2022}. The most commonly used data augmentation technique is to apply random shifts to the pixel observations. In Figure~\ref{fig:figure_SI_rl_algos}, we compare the performance of DrM and SAC with and without random shift augmentation using an identical CNN encoder structure (see Methods section of the main text). For both algorithms, we find that data augmentation significantly helps to achieve efficient learning and high performance. Furthermore, we observe that DrM training is more stable, resulting in higher episodic rewards and with that reduced average episode lengths --- see Figure~\ref{fig:figure_SI_rl_algos}(b,c). 

\section{Reward functions}

\begin{figure}[h!]
\centering
\includegraphics[width=0.8\textwidth]{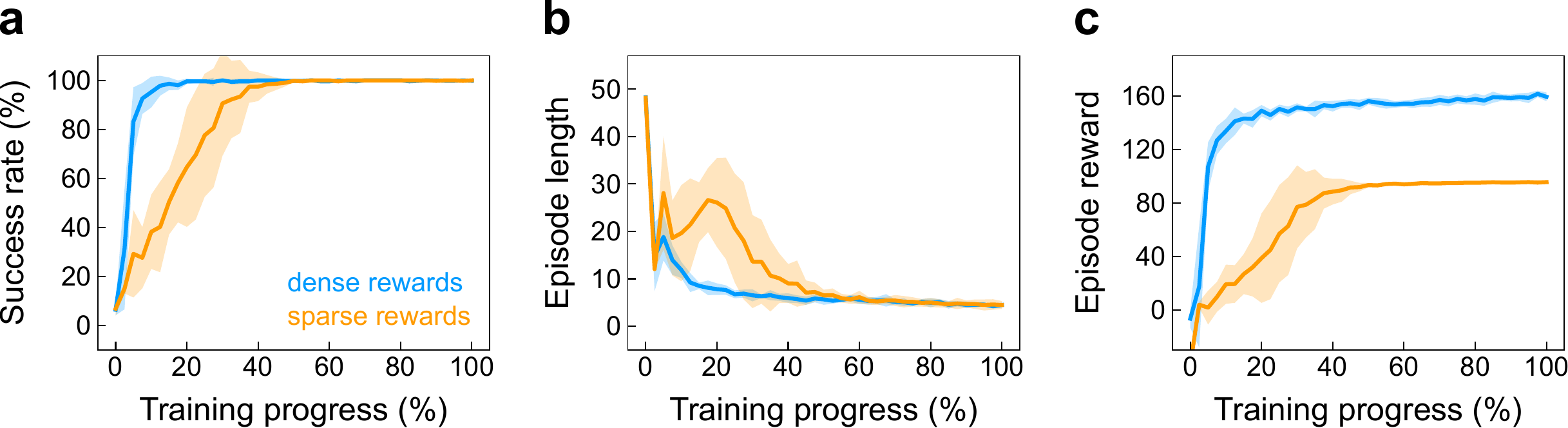}
\caption{\textbf{Comparison of different reward functions.} (a-c) 
Training curves for a cubic system as a function of percentage training cycles for dense and sparse reward functions as indicated. Solid lines represent the mean, while shaded areas indicate the 95\% confidence interval calculated from five training runs with different random seeds.}
\label{fig:figure_SI_sparse_rewards_draft}
\end{figure}

The appropriate choice and design of the reward function is a crucial component in enabling the agent to learn efficiently. 
In this work, we made use of a dense reward scheme, discounted by the step number $t$ within the episode (see Methods section of the main text). In Figure~\ref{fig:figure_SI_sparse_rewards_draft}, we compare the dense reward scheme with sparse rewards where each step yields a constant reward (penalty) of -1. Like in the dense reward scheme, the agent receives a final reward of +100 for reaching the target orientation.
In this case, the agent does not receive any rewards based on the direction and magnitude of his actions. 
Using a sparse-reward setting, the agent's training performance is slowed down compared to a dense-reward scheme. Nevertheless, the agent is still able to achieve  the same performance compared to the dense reward scheme towards the end of the training.

\clearpage

\section{Agent generalization to different space groups}

During the  training stage, the reinforcement-learning (RL) agent is provided with sequences of Laue patterns of a specific crystal structure and space group. 
To investigate the achieved agent generalization, we evaluate  an agent, trained on a single space group, to other space groups within the same crystal structure. The space group provides information about allowed and forbidden $(hkl)$ reflections, meaning Laue patterns of crystals with different space groups can vary significantly due to different selection rules. For example, in the simple cubic space group 221, all $(hkl)$ combinations are allowed, while in the face-centered cubic space group 225 only those combinations with either all even or all odd sums of $(hkl)$ are allowed.\\

\begin{figure}[h!]
\centering
\includegraphics[width=0.95\textwidth]{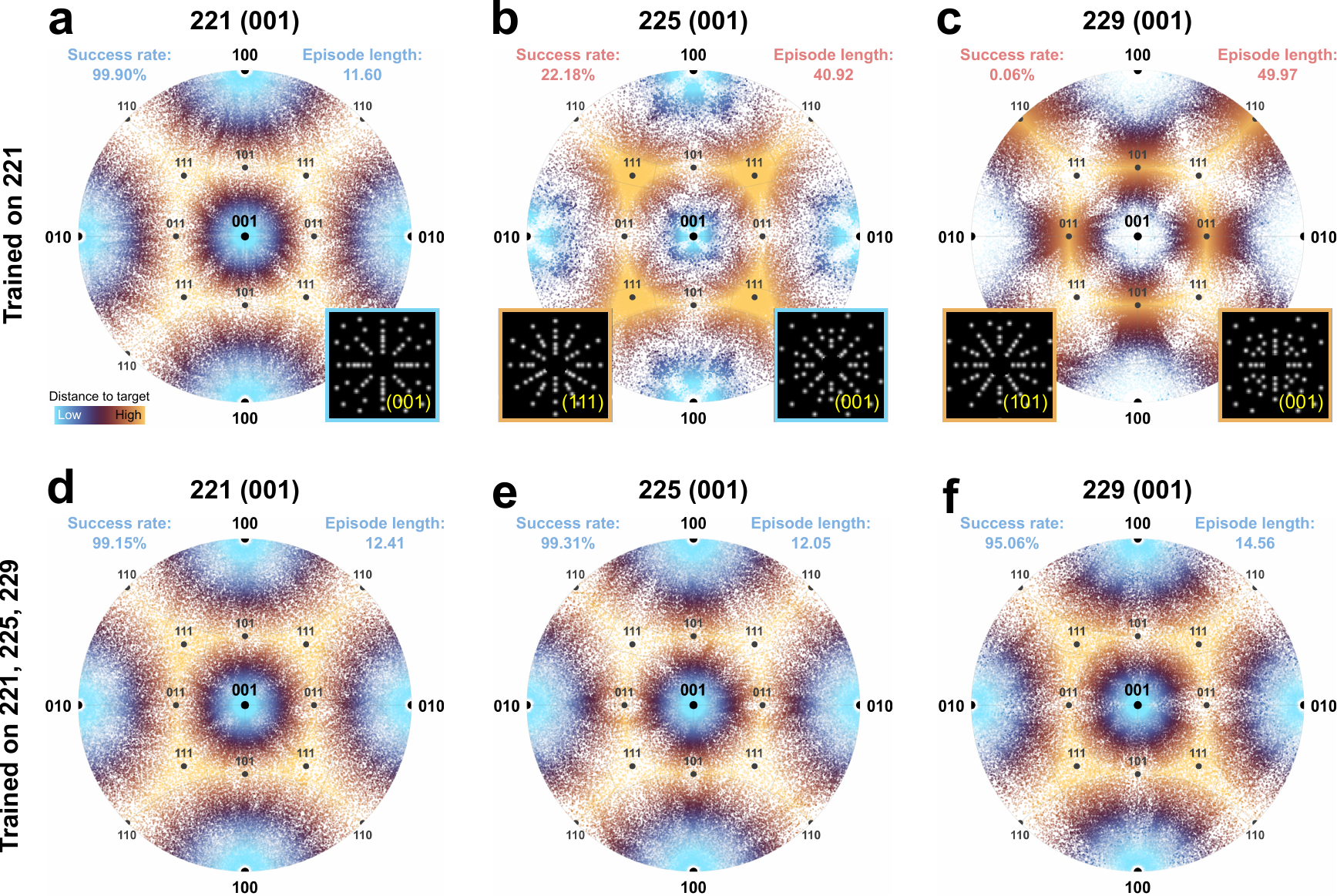}
\caption{\textbf{Agent generalization to different cubic space groups}. (a-c) Laue agent, trained to align a crystal with simple cubic (221) space group along the (001) high-symmetry orientation, evaluated on simple cubic (221), face-centered cubic (255), and body-centered cubic (229) crystal structures for 10000 episodes using stereographic projections. A high performance can be observed for identifying the (001) orientation in space group 221. For space groups 225 and 229 the agent seems to be directed towards the (111)- and (101)-style directions respectively. Examples of high-symmetry Laue patterns are shown at the bottom of each stereographic projection. (d-f) Laue agent, trained to align a crystal with any of the space groups (221, 225,  229) along the (001) high-symmetry orientation, evaluated on  space group 221, 255, and  229 cubic crystal structures for 10000 episodes. In this case, the training produced a robust agent for different cubic crystal structures. To better illustrate the agent's behavior, the step size is limited to two degrees.}
\label{fig:figure_SI_cubic_space_groups}
\end{figure}

\begin{figure}[h!]
\centering
\includegraphics[width=0.8\textwidth]{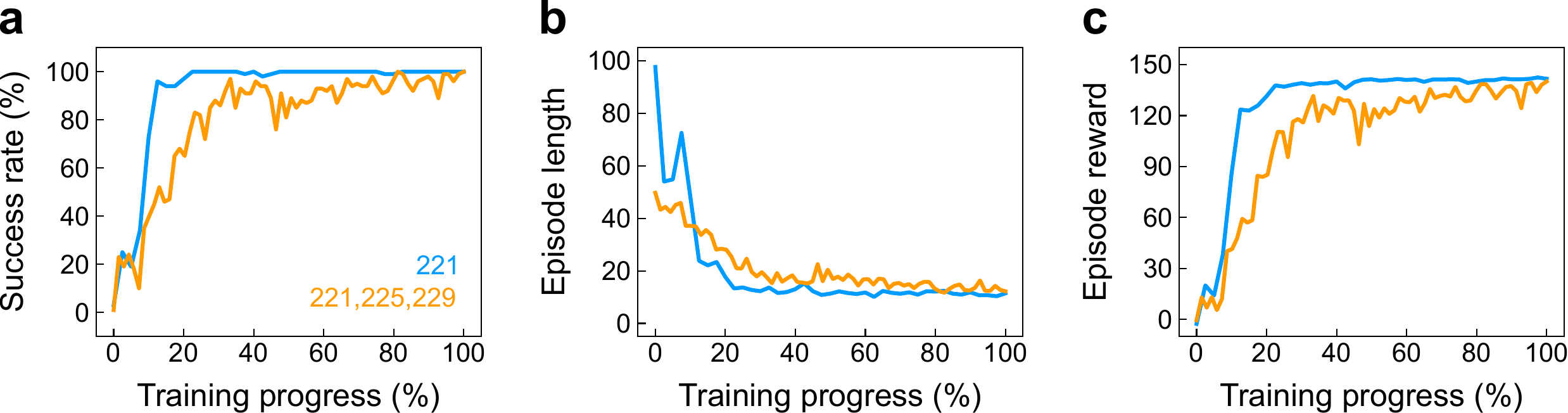}
\caption{\textbf{Agent generalization to different cubic space groups}. (a-c) Training curves for a single seed showing success rate, episode length, and success rate for training an agent on either simple cubic (221) or a mixture of simple cubic, face-centered cubic, and body-centered cubic crystal structures (221,225,229). The training with purely simple cubic crystal structure (mixture model) was performed over 200000 (400000) training steps.}
\label{fig:figure_SI_cubic_space_groups_training_curves}
\end{figure}

In Figure~\ref{fig:figure_SI_cubic_space_groups}, we demonstrate the agent's ability to transfer its learned behavior to space groups different from its training space group. We initially trained an RL agent only to identify the (001) high-symmetry direction for simple cubic 221. As expected, the agent performs well when evaluated on space group~221~(Figure~\ref{fig:figure_SI_cubic_space_groups}a). In a next step, the agent is evaluated on different cubic space groups such as face-centered cubic~225~(Figure~\ref{fig:figure_SI_cubic_space_groups}b) or body-centered cubic~229~(Figure~\ref{fig:figure_SI_cubic_space_groups}c). In those cases one observes that the agent is targeting  different high-symmetry orientations in the majority of cases. This indicates that the agent has learned intricate relations between high-symmetry lines leading to a (001)-style orientation for the simple cubic crystal structure. For example, when evaluating an agent trained on 
space group 221 and evaluating it on 
space group 225, the agent primarily targets either (111) or (001) high-symmetry points with probabilities of roughly 78\% and 22\% respectively. If evaluating on 
space group 229, the agent seems to strongly prefer the (101)-style orientations instead with close to 100\% probability. To increase the agent generality in succeeding to reach the (001) orientation for different cubic space groups, we conduct training runs with additional domain randomization in terms of 
randomly selecting one of the (221, 225, 229) space groups at the beginning of each training episode. 
The performance of the trained agent when evaluated on individual space groups is visualized in Figure~\ref{fig:figure_SI_cubic_space_groups}(d-f). It is evident that the resulting agent has learned to align on the (001) orientation independent of the space group. In Figure~\ref{fig:figure_SI_cubic_space_groups_training_curves}(a-c) we show the respective training curves. 

\section{Experimental control setup}

\begin{figure}[h!]
\centering
\includegraphics[width=0.6\textwidth]{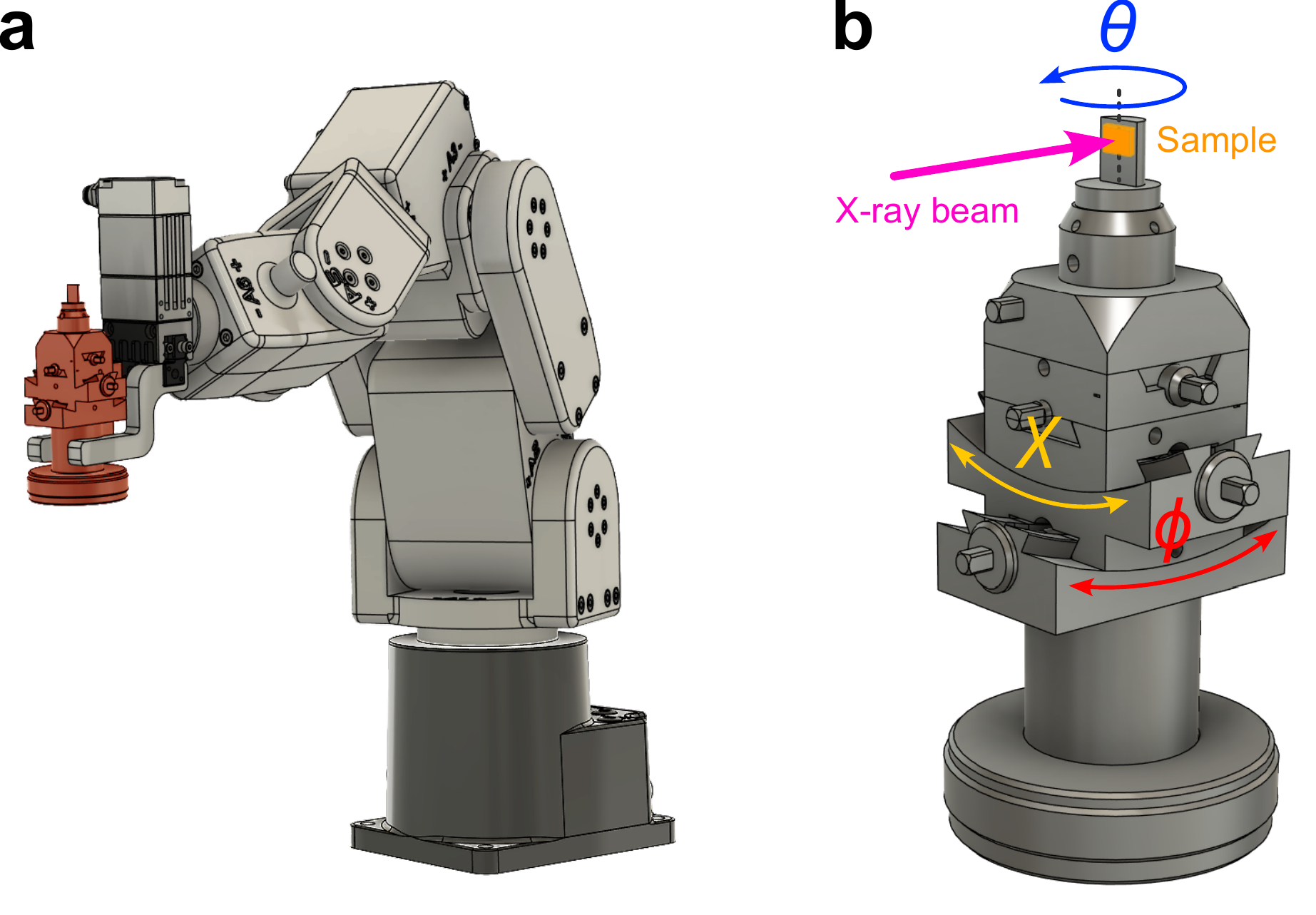}
\caption{\textbf{Experimental control setup}. (a) Mecademic Meca500 six-axis robot arm, equipped with a MEGP 25E end effector and custom aluminium grippers. The crystal sample is mounted on a goniometer head (highlighted in red and enlarged in (b)). The sample center of rotation is manually positioned in front of the x-ray source before the angular alignment is initiated. The robotic control commands are executed using the Mecademic Python API. (b) Conventional goniometer head (model 1002) from Huber with indicated angles $(\theta,\chi,\phi)$, used to create the initial angular offsets of the sample as described in the Methods section of the main text.}
\label{fig:figure_SI_robot_arm}
\end{figure}

\clearpage

\printbibliography[title={Supplementary References}]